\def\year{2022}\relax
\documentclass[letterpaper]{article} 
\usepackage{aaai22}  
\usepackage{times}  
\usepackage{helvet}  
\usepackage{courier}  
\usepackage[hyphens]{url}  
\usepackage{graphicx} 
\usepackage{natbib}  
\usepackage{caption} 
\DeclareCaptionStyle{ruled}{labelfont=normalfont,labelsep=colon,strut=off} 
\usepackage{algorithm}
\usepackage{array}
\usepackage{algorithmic}
\usepackage{amsmath}
\usepackage{amsfonts}
\newcommand{\xtrn}{\ensuremath{\mathcal{X}_{\mathrm{trn}}}}

\usepackage{newfloat}
\usepackage{listings}
\floatstyle{ruled}
\newfloat{listing}{tb}{lst}{}
\floatname{listing}{Listing}
\title{Heuristic Search Planning with Deep Neural Networks
using Imitation, Attention and Curriculum Learning}
\author {
     Leah Chrestien,
     Tomáš Pevný,
     Antonín Komenda,
     Stefan Edelkamp
}
\affiliations {
 Department of Computer Science, Faculty of Electrical Engineering \\
Czech Technical University in Prague \\
 chreslea@fel.cvut.cz, pevnytom@fel.cvut.cz, antonin.komenda@fel.cvut.cz, stefan.edelkamp@fel.cvut.cz
}
\usepackage{bibentry}

\begin{document}

\maketitle

\begin{abstract}
Learning a well-informed heuristic function for hard task planning domains is an elusive problem. 
Although there are known neural network architectures to represent such heuristic knowledge, it is not obvious what concrete information is learned and whether techniques aimed at understanding the structure help in improving the quality of the heuristics.

This paper presents a network model to learn a heuristic capable of relating distant parts of the state space via optimal plan imitation using the attention mechanism, which drastically improves the learning of a good heuristic function. To counter the limitation of the method in the creation of problems of increasing difficulty, we demonstrate the use of curriculum learning, where newly solved problem instances are added to the training set, which, in turn, helps to solve problems of higher complexities and far exceeds the performances of all existing baselines including classical planning heuristics. We demonstrate its effectiveness for grid-type PDDL domains. 
\end{abstract}

\section{Introduction}
Classical Planning has always relied on strong heuristic functions to approximate distances to the nearest goal \cite{bonet2001planning}. Generally speaking, the quality of heuristics is measured by how well it performs when used inside a planner, i.e., it depends on the quality of the solution and the time taken to generate it. A major drawback of classical planning is the need to formulate problems by extensively capturing information from the environment. Recent years observe a progress in using visual representations to capture the specifics of a problem \cite{asai2017classical}. Yet, there is still a big gap between the length of optimal plans and the plans found by planners using learnt heuristic functions.

A significant amount of importance is given to developing deep networks that are able to learn strong heuristics \cite{ernandes2004likely} and policies \cite{torrey2006skill}. This is particularly common in Reinforcement Learning which uses positive and negative feedback to learn the correct sequence of next actions. In learning for planning, there is no such provision; instead, it relies heavily on either hand-coded logical problem representations \cite{yoon2012inductive} or deep convolution neural networks \cite{groshev2017learning} that imitate an expert. While there exists successful approaches in training neural networks to learn heuristic estimates of various problem domains \cite{arfaee2011learning,us2013learning,groshev2017learning}, designing a meaningful neural network architecture to extract the relevant information from the data set is still an open-ended problem. 
This work extends the work 
by~\citeauthor{schaal1999imitation} 
(\citeyear{schaal1999imitation}) 
and~\citeauthor{groshev2017learning} (\citeyear{groshev2017learning}) 
by addressing limitations of convolutional neural network and by using
imitation and curriculum learning to train from plans of more difficult problem instances for hard planning domains. Specifically, we propose to use self-attention and position encoding~\cite{vaswani2017attention}, as we believe a strong heuristic function needs to relate "distant" parts of the state space. 

In our default experimental settings, neural networks (NNs) realizing heuristic functions are trained on plans of small problem instances created by classical planners. While this allows us to generalize across more difficult instances (some of which are still solvable by classical planners) such that we can measure distances to optimal plan lengths, they do not achieve the best results for two reasons. 
First, even though the generalization of A*-NN is surprisingly good as will be seen below, there is still a scope of large-scale improvement on previously unseen, larger and more complex environments.
Second, classical domain independent planners can solve only small problem instances anyway, which means that obtaining plans from large ones is difficult. We demonstrate that this problem can be partially mitigated by curriculum learning, where the NN is retrained / fine-tuned using plans from problems it has previously solved.

The proposed approach is compared to state-of-the-art domain-independent planners, namely SymBA*\cite{torralba2014symba}, Lama \cite{richter2010lama}, and Mercury \cite{katz2014mercury} and to currently best combination of A* and CNNs~\cite{groshev2017learning} on three grid domains: (1) \textbf{Sokoban} where each maze consists of walls, empty spaces, boxes, targets and an agent; the goal is to push the boxes to target locations; the boxes can only be pushed and not pulled in the game; 
(2) \textbf{Maze-with-Teleports} where the goal for an agent is to reach the goal position via interconnected teleports;
(3) \textbf{Floor-Tile} where the tiles in a given maze are to be colored alternatively by agents (in our case, two agents) that are assigned a particular color each. At no stage can an agent step on a tile that has been colored. The approach is domain-independent and only requires the selection of propositions in the PDDL file for spanning the underlying grid and the objects moving on it. It uses either policy or heuristic value heads.

We have chosen these domains because (1) all can be easily translated into an image based representation on which we can apply the convolution operation;\footnote{If the domain can be described by graph, we can replace image convolution by graph-convolution, though we do expect the experimental results on these domains to exhibit very different behavior.} (2) Sokoban is PSPACE-complete~\cite{Culberson199965} and known to be a challenging problem in Deep Learning for Planning \cite{fern2011the}; therefore, an improvement in Sokoban's policy and heuristic learning implies success in other two dimensional planning domains; (3) Maze-with-Teleports have non-local actions as the maze agent is being teleported to different parts of the maze via teleports; (4) Floor-Tile is NP-Complete and allows easy generation of mazes of increasing difficulty, which makes it a great choice of domain for improving the learnt heuristics via curriculum learning \cite{bengio2009curriculum}. 

The paper is organised as follows. We first discussed existing relevant research that has been carried out in deep learning for planning, especially in image based games. Next, we explore the formal basics of classical planning. Then, we highlight the shortcomings of a prior state of the art and propose a solution that addresses some of these shortcomings. Here, we introduce the basics of the attention mechanism from NLP and explain the role of positional encoding in learning distances. Then, the proposed networks are compared to other state of the art methods that attempt to solve Sokoban, Floor-Tile and Maze-with-Teleports. Finally, we conclude with an overall synopsis of our work and lay the ground for future work. 

\section{Related Work}

The implementation of neural networks (NN) in learning policies and heuristics for deducing strategic positions or moves in various game domains has been studied extensively in the past. In Chess \cite{thrun1994learning}, NNs have been used to evaluate chess board functions from the outcome of various games. A combination of supervised learning and reinforcement learning has been implemented in games such as Go \cite{silver2017mastering}, which uses value networks to evaluate board positions and policy networks that choose successive actions. In Backgammon \cite{tesauro2002programming}, temporal difference learning was employed to train networks from millions of gameplay. Even single agent games such as Sokoban \cite{racaniere2017imagination} and Rubik's cube \cite{agostinelli2019solving} use NNs to solve the respective puzzles. The relative success of NNs in reinforcement learning has led researchers to believe that a similar success may be achieved in learning heuristic functions for various classical planning domains.

\citeauthor{fern2011the} (\citeyear{fern2011the}) 
presented the earliest known work that combines planning and learning and concludes that the performance of NNs is promising in learning heuristic functions. The work of \cite{us2013learning} attempts to combine classical planning and deep learning by modifying costs and learning an improved heuristic function that generates good quality plans.  \citeauthor{arfaee2011learning}
(\citeyear{arfaee2011learning}) generated strong heuristics by repeatedly training on a set of weak heuristics by the boot-strapping procedure which are then used inside a classical planner. Neural search policies \cite{gomoluch2020learning} rely on parameter learning to generate heuristics during the actual search process. The learning procedure in \cite{srivastava2011new} leads to generalizations of classical plans by identifying landmark actions that may be repeatedly applied during learning.
Delphi applies deep learning on performance profile graphs to choose the cost-optimal planner in a  portfolio~\cite{Sievers0SSF19}


\citeauthor{asai2017classical}~\shortcite{asai2017classical} used classical planning to generate data and designs in an architecture 
that returns a visualised plan execution. In \cite{groshev2017learning}'s work, the training samples were generated by classical planning, and imitation learning \cite{schaal1999imitation} was performed on this data set for policy and heuristic learning. Unlike \citeauthor{fikes1971strips}~\shortcite{fikes1971strips} and \citeauthor{shavlik1989acquiring}~\shortcite{shavlik1989acquiring} that used hand-coded domains to represent problems, \citeauthor{asai2017classical}~\shortcite{asai2017classical} and \citeauthor{groshev2017learning}~\shortcite{groshev2017learning} learned useful information from the input data through image-based state descriptions.

Our approach significantly differs from prior work in planning and learning: (1) we do not require any pre-designed model of the problem domain or the state transition system; instead our model learns primarily from optimal plans; (2) our model uses attention mechanism with positional encoding designed to learn distances in the heuristic network; this generates near optimal solutions for complex problem instances where other techniques often fail. 


\section{Classical Planning}
We construct our problem domains in a classical setting, i.e. fully observable and deterministic.

In classical planning, a STRIPS~\cite{fikes1971strips} planning task is defined by a tuple $\Pi = \langle F, A, I, G \rangle$. $F$ denotes a set of facts which can hold in the environment (for instance, in Sokoban, a particular box at a particular position is a fact). A state $s$ of the environment is defined as a set of facts holding in that particular $s$, i.e. $s \subseteq F$. The set of all states is, therefore, defined as all possible subsets of $F$ as $S=2^F$. $I \in S$ is the initial state of the problem and $G \subseteq F$ is a goal condition comprising facts which has to hold in a goal state. An action $a$, if applicable and applied, transforms a state $s$ into a successor state $s'$ denoted as $a(s) = s'$ (if the action is not applicable, we assume it returns a distinct failure value $a(s) = \bot$). All actions of the problem are contained in the action set $A$, i.e. $a \in A$. The sets $S$ and $A$ define the state-action transition system. 

Let $\pi=({a}_1, {a}_2, \ldots, {a}_l)$, we call $\pi$ a plan of length $l$ solving a planning task $\Pi$ iff $a_l( \ldots a_2(a_1(I)) \ldots ) \supseteq G$. We assume a unit cost for all actions, therefore the plan length and plan cost are equal. Moreover, let $\pi_s$ denote a plan from a state $s$, not $I$. An optimal solution (plan) is defined as a minimal length solution of a problem $\Pi$ and is denoted as $\pi^*$ together with its length $l^*=|\pi^*|$.

A heuristic function $h$ is defined as $h:S \rightarrow \mathbb{R}^{\ge 0}$ and provides an approximation of the optimal plan length from a state $s$ to a goal state $s_g \supseteq G$, formally $h(s) \approx l^*$, where $l^* = |\pi^*_s|$.

In our experiments, we choose domains encoded in PDDL \cite{fox2003pddl2}, where a planning problem is compactly represented in a lifted form based on predicates and operators. This representation is grounded into a STRIPS planning task $\Pi$, which is subsequently solved by the planner using a heuristic search navigating in the state-action transition system graph and resulting in a solution plan $\pi$. 

\section{Planner's Architecture}

To learn a heuristic function for a planning domain 
that estimates the cost-to-go in a current state is one of the holy grails in AI~\cite{thebook,Mostow1989DiscoveringAH}. 
 
In our approach we combine a domain-independent
planner with a trained neural networks heuristic for its improved guidance during the overall search. This way we are bridging symbolic and sub-symbolic reasoning, given that neural networks are data-driven, and task planning requires 
symbolic reasoning in some form of logical calculus. 

To bridge this gap, we highlight that any (deterministic) plan is a sequence of actions, but also on states. 
Given the initial state $I$, each partial plan 
$\pi=({a}_1, {a}_2, \ldots, {a}_k)$, $k<l$ 
induces a sequence of states $({s}_0 = I, {s}_1, {s}_2, \ldots, {s}_k)$ with $s_k = a_k( \ldots a_2(a_1(I))$.
Extracting the state sequence was not difficult to program, on could also use off-the-shelf tools like VAL \cite{HoweyLF04}. 

That a state is propositional is not a limitation to the approach. We might also uses finite-domain variables as in SAS$^+$~\cite{BackstromN95}, or include real-valued variables for metric planning~\cite{fox2003pddl2}. 

Any encoded state is an input for the network. In our setting we use a \emph{one-hot bitvector encoding} 
of the propositions.
The output of the network is the heuristic value called \emph{value head}, in some cases, together with a distribution of action to take, called \emph{policy head}. 
We prefer optimal plans for selecting training instances for the neural network, which might be generated by any optimal planner. More precisely, given the plans in the training set, with imitation learning we generate 
pairs $(s_i,\delta(s_i))$, where $\delta$ is cost of an optimal plan from $s_i$ to the goal. For the sake of simplicity, $\delta(s_i)$ is the distance  $l-i$ of the state $s_i$ to the goal $s_l$ in the optimal plan $({s}_0 = I, {s}_1, {s}_2, \ldots, s_i, \ldots ,{s}_l)$. Evaluating the network for a given state, directly serves as estimator in our heuristic search planner. With curriculum learning we also include newly found close-to-optimal plans to retran the NN learner.

In some cases we will also take pairs of $(s_{i-1},a_i)$ to train the network for its policy head.  Since our aim is close-to-optimal plans, as the search algorithm for exploring the planning state space, we employed A*~\cite{hart1968formal}.
For training the network we run the known backpropagation algorithm on input batches together with stochastic gradient decent~\cite{BottouB07} to minimize the network error, that is computed with a simple loss function applied to the predicted and real value. Once trained, the heuristic estimate can be extracted very efficiently for each state from the value head. In some cases action selection can be based on the policy head.

As we use optimal plans, we train the perfect heuristic estimate, but even for considerably good plans, a goal estimate can be defined on other plans that have been generate. All options that have
been suggested also work for solutions of simplified planning task, such as abstraction, relaxation can be used to generate a set of plans from which we can extract distances.
While we assume to be able to identify the propositions 
that generate an underlying grid, we emphasize that the above architecture is general and could be applied to any state/action pairs extracted from the selection of good or optimal plans. We take a rather generic network model, but with key ingredients of position attention convolution, trained with imitation and curriculum learning.

\section{Learning the Heuristic}
\begin{figure*}[t]
\centering
\includegraphics[width=\textwidth]{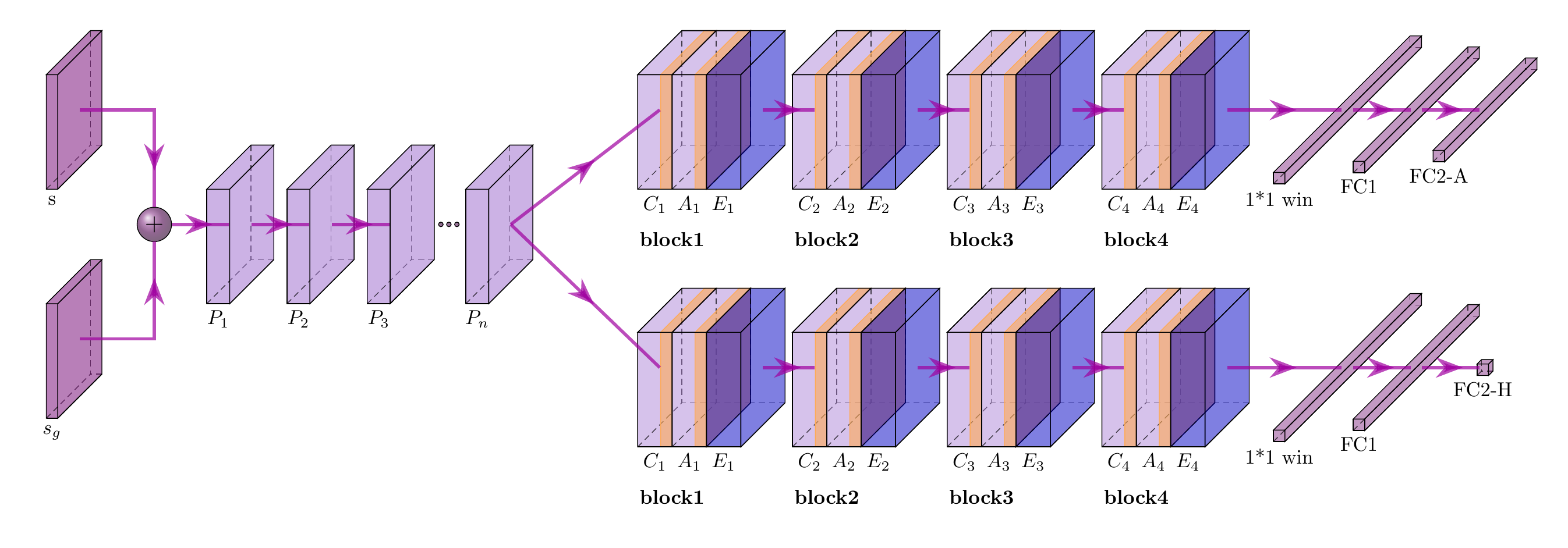}
\caption{The structure of our neural network. A current state $s$ and a goal state $s_g$ are fed into a variable number of pre-processing convolution (pre-conv) layers, $P_{1}..P_{n}$. All convolution filters in the pre-conv layers are of the same shape $3\times3$ with 64 filters. Then the network splits into two branches and each branch has four blocks, each block containing a convolution layer (C) followed by a multi head attention operation (A) and a positional encoding layer (E). There are 180 filters in each of these convolution layers in the blocks. At all stages, the original dimension of the input is preserved. The output from block 4 is flattened by applying a 1 × 1 window around the agent's location before being passed onto the fully connected layers (FC1) and the action prediction output (FC2-A) with 8 actions (move: up, down, left, right and push: up, down, left, right) and a single output for heuristic prediction (FC2-H).
For the sake of picture clarity, skip connections are not shown in the neural network.}
\label{fig:neural_network}
\end{figure*}

This section describes the proposed neural network for planning in maze-like PDDL domains. But before, we introduce the notation and briefly discuss the state-of-the-art along with the proposed modification. In the end, we discuss the concrete architecture 
we used for all the benchmark domains (Sokoban, Floor-Tile and Maze-with-Teleports). We extract the grid layout automatically, but for the time being, we assume to have the grid dimensions
$h$ and $w$ available to define the network.

\subsection{Formal notations for the proposed neural networks}
The input to the neural network is denoted as $\vec{x} \in \mathbb{R}^{h,w,d_0},$ where $h$ and $w$ is the height and width of the maze respectively, and $d_0$ varies with the number of channels as explained above. Intermediate outputs are denoted by $\vec{z}^{i} = L_i(\vec{z}^{i-1}),$ where $L$ is some neural network layer (convolution $C$, attention $A$, or position encoding $E$) and for the sake of convenience, we set $\vec{z}^0 = \vec{x}.$ All $\vec{z}^{\cdot}$ are three dimensional tensors, i.e. $\vec{z}^i \in \mathbb{R}^{h,w,d_i}.$ Notice that all intermediate outputs $\vec{z}^i$ have the same width and height as the maze (ensured by padding), while the third dimension which is the number of output filter(s) differs. Value $\vec{z}^{i}_{u,v}$ denotes a vector created from $\vec{z}^i$ as $(\vec{z}^{i}_{u,v,1},\vec{z}^{i}_{u,v,2},\ldots, \vec{z}^{i}_{u,v,d_i}).$ Below, this vector will be called a \textit{hidden vector} at position $(u,v)$ and can be seen as some description of the properties of this position.  

\subsection{The proposed neural network}
The best published NN implementing a heuristic function for Sokoban was proposed in~\cite{groshev2017learning}. The network's shape resembled letter Y, as it has two heads, and it contains only convolution layers. The first seven convolution layers were shared (we call them pre-conv layer abbreviating preprocessing-convolution). Then, the network splits to yield two sets of outputs: (i) the estimate of the heuristic function and (ii) the policy. After the split, each path to the output contained seven convolution layers followed by two dense layers. Although the heuristic function should be sufficient for planning purposes, ~\cite{groshev2017learning} states that training the network to estimate the policy helps in computing a better 
heuristic function.

Our criticism of the architecture is that convolution is strictly a local operator. This means that the hidden vector $z^{i+1}_{u,v,\cdot}$ is calculated from hidden vectors $\{z^{i}_{u',v',\cdot} | u' \in \{u-1,u,u+1\}, v' \in \{v-1,v,v+1\}\},$ where we assume convolution to have dimension $3 \times 3$ as in~\cite{groshev2017learning}. This limits the neural network in synthesizing information from two distant parts of the maze. Yet, we believe that any good heuristic requires such features, since Sokoban, Floor-Tile and Maze-with-Teleports are non-local problems.

\section{Convolution,  Attention,  and  Position  Encoding}

Therefore, we turn our attention to self-attention mechanism,~\cite{vaswani2017attention} first introduced in NLP, as it allows to relate distant parts of input together. The output of self-attention from  $z^{i}$ is calculated in the following manner. At first, the output from previous layer $z^{i}$ is divided into three tensors of the same height, width, and depth, i.e.
\begin{alignat*}{2}
\vec{k} & = z^i_{\cdot, \cdot, j}\quad j \in \left\{1,\ldots, \frac{d_i}{3}\right\}\\
\vec{q} & = z^i_{\cdot, \cdot, j}\quad j \in \left\{\frac{d_i}{3}+1,\ldots, \frac{2d_i}{3}\right\} \\
\vec{v} & = z^i_{\cdot, \cdot, j}\quad j \in \left\{\frac{2d_i}{3}+1,\ldots, d_i\right\} 
\end{alignat*}
then, the output $z^{i+1}$ at position $(u,v)$ is calculated as
\begin{equation}
\vec{z}^{i+1}_{u,v} = \sum_{r=1,s=1}^{h,w} \frac{\exp(\vec{q}_{u,v} \cdot \vec{k}_{r,s} )}{\sum_{r'=1,s'=1}^{h,w} \exp(\vec{q}_{u,v} \cdot \vec{k}_{r',s'})} \cdot v_{r,s}
\label{eq:self-attention}
\end{equation}
Self attention, therefore, makes a hidden vector $z^{j+1}_{u,v}$ dependent on all hidden vectors $\{z^{j}_{r,s} |r\in\{1,\dots,h\}, s \in \{1,\ldots,w\}\}$, which is aligned with our intention.  The self-attention also preserves the size of the maze. A multi-head variant of self-attention means that $z^{i}$ is split along the third dimension in multiple $\vec{k}$s, $\vec{q}$s, and $\vec{v}$s. The weighted sum is performed independent of each triple $(k,q,z)$ and the resulting tensors are concatenated along the third dimension. We refer the reader for further details to~\cite{vaswani2017attention}.

While self-attention captures information from different parts of the maze, it does not have a sense of a distance. This implies that it cannot distinguish close and far neighborhoods. To address this issue, we add positional encoding, which augments the tensor $z^{i} \in \mathbb{R}^{h,w,d_i}$ with another tensor $\vec{e}\in \mathbb{R}^{w,h,d_e}$ containing outputs of harmonic functions along the third dimension. Harmonic functions were chosen, because of their linear composability properties \cite{vaswani2017attention}.
\footnote{The composability of harmonic functions is based on the following property $\cos(\theta_1 + \theta_2) = \cos(\theta_1)\cos(\theta_2) - \sin(\theta_1)\sin(\theta_2) = (\cos(\theta_1), \sin(\theta_1)) \cdot (\sin(\theta_1), \sin(\theta_2)),$ where $\cdot$ denotes the inner product of two vectors, which appears in Equation~\eqref{eq:self-attention} in inner product of $\vec{q}_{uv}$ and $\vec{k}_{rs}$.}
Because our mazes are two dimensional, the distances are split up into row and column distances where $p, q \in [0, d_{i}/4)$ assigns positions with sine values at even indexes and cosine values at odd indexes. The  positional encoding tensor $\vec{e}\in \mathbb{R}^{w,h,d_e}$ has elements equal~to
\begin{alignat*}{4}
\vec{e}_{u,v,2p}       &= \sin\left(\theta(p) u\right)&
\vec{e}_{u,v,2p+1}     &= \cos\left(\theta(p)  u\right)\\
\vec{e}_{u,v,2q+\frac{d_e}{2}}  &= \sin\left(\theta(q) v\right)\quad &
\vec{e}_{u,v,2q+1+\frac{d_e}{2}} &= \cos\left(\theta(q) v\right),
\end{alignat*}
where $\theta(p) = \frac{1}{10000^{\frac{4p}{d_{e}}}}$.  On appending this tensor to the input $z^i$ along the third dimension, we get 
$$z^{i+1}_{u,v,\cdot} = [z^i_{u,v,\cdot}, e_{u,v,\cdot}].$$

With respect to the above, we propose using blocks combining Convolution, Attention, and Position encoding, in this order (we call them CoAt blocks), as a part of our NN architecture. The CoAt blocks can therefore relate hidden vectors from a local neighborhood through convolution, from a distant part of the maze through attention, and calculate distances between them through position encoding, as has been explained in~\cite{tsai2019transformer}. Since CoAt blocks preserve the size of the maze,\footnote{Convolution layers are appropriately padded to preserve sizes.} they are "scale-free" in the sense that they can be used on a maze of any size. Yet, we need to provide an output of a fixed dimension to estimate the heuristic function and the policy. The outputs of the last CoAt blocks are flattened by a $1 \times 1$ window, centered around the agent's location and fed to a fully-connected layer and an output head (see Figure \ref{fig:neural_network}). For example, assuming $z^L$ to be the very last layer, and agent is on position $u,v$, the vector $z_{u,v,\cdot}$ is of constant dimension equal to the number of filters and is used as an input to the fully-connected layers providing the desired outputs (heuristic values, policy).

Next, we describe the implementation of CoAt blocks in the network architectures we used for all the domains.
The network 
is shown in Figure~\ref{fig:neural_network} and its structure is similar to that of~\cite{groshev2017learning}. It uses preprocessing convolution layers $P_1,\ldots,P_n,$ $n = 7,$ (further called \textit{pre-conv}) containing 64 filters where each convolution filter is of the shape $3\times3$; after the network splits into two heads, it uses four CoAt blocks in each head instead of seven convolution layers used in~\cite{groshev2017learning}. The convolution layers in the CoAt blocks contain 180 filters of size $3 \times 3$ each. The attention block uses two attention heads. Each head is finished by two fully-connected layers with reduction to a fixed dimension as described above.

The input to the network is the current state of the game and a goal state, $s$ and $s_{g}$, respectively.
Each state is represented by a tensor of dimensions equal to width and height (fixed to $10 \times 10$ for Sokoban, to $15 \times 15$ for Maze-with-Teleorts, and to $4 \times 4$ for Floor-Tile) of the maze~$\times$~objects. The objects stands for one-hot encoding of the object states on a grid position (e.g., for Sokoban, we have wall, empty, box, agent and box target, for Maze-with-Teleport agent, wall, floor, goal and teleports 1-4, and for Floor-Tile
agent1, agent2, black and white), 
which we could derive automatically from the grounded representation. An important design detail is that all convolutions are padded, which means that the output has the same dimension as the input, and they feature skip-connections\footnote{A layer $c$ augmented by a skip connection calculates the output as $x + c(x)$ instead of the usual $x$.} alleviating a possible vanishing gradient~\cite{he2016deep}. 

We have implemented two different versions of the network according to their heads: dual-head with estimating policy and heuristic value and single-head with just estimating heuristic value.. 
In Maze-with-Teleports and Sokoban the dual-head representation performed best, while the best network for Floor-tile uses just a single head estimating heuristic value (which means there is no separate head estimating policy). 
The presence of two agents would make it difficult to design a policy network in a domain-independent setting and would result in a much larger network, which is inconvenient, time-consuming and computationally expensive. The other implementation details of the Floor-Tile network are the same as the heuristic networks of Sokoban and Maze-with-Teleports (see Figure \ref{fig:neural_network}).

\section{Imitation and Curriculum Learning}

Imitation learning~\cite{pomerleau1989alvinn} is a framework for learning a behavior policy from demonstrations. We present demonstrations in the form of optimal state-action plans, with each pair indicating the action to take at the state being visited.
Generally, imitation learning is useful when it is easier for an expert to demonstrate the desired behaviour rather than to specify a reward function which would generate the same behaviour or to directly learn the policy

A curriculum refers to an interactive system of instruction and learning with specific goals, contents, strategies, measurement, and resources. The desired outcome of curriculum is a successful transfer and/or development of knowledge, skills, and attitudes.
In the context of AI, curriculum learning is a way of training a machine learning model where more difficult aspects of a problem are gradually introduced in such a way that the model is always optimally challenged. 

Curriculum learning~\cite{ELMAN199371} describes a type of learning in which we first start out with only easy examples of a task and then gradually increase the task difficulty. Humans have been learning according to this principle ever since, but in the common learning setting, we use neural networks and instead let them train on the whole data set.

Curriculum learning strategies have been successfully employed in different areas of machine learning, for a wider range of tasks~\cite{bengio2009curriculum}. However, the necessity of finding a way to rank the samples from easy to hard, as well as the right pacing function for introducing more difficult data can limit the usage of the curriculum approaches. 

 In order to extend the training set without providing any additional plans that the neural network would not be able to solve, we turn our attention to a form of curriculum learning for neural networks. 
This approach that partially circumvents this problem by re-training from unseen test samples of increasing complexity. 

In our case, curriculum learning is used to develop scale-free heuristic values for a wider selection of AI planning problems. Specifically, in our experiments we have quickly reached the capability of  planners at larger sizes. To further improve our heuristic function to scale to bigger problems, we re-train our network by extending the training set to include harder problem instances.

We first train the heuristic network on a training set containing easy problem instances quickly solvable by an optimal, then use this NN as a heuristic function inside A*, and then extend the training set by more difficult problem instance the NN has solved and retrain the NN. Thus, we perform a bootstrap, where the NN is gradually trained on more difficult problem instances.

This way, curriculum learning plays an important role in improving the performance of the heuristic network on not just the trained dimensions but also on higher dimensions by extrapolation. 
For curriculum learning, the learning rate is reduced in successive training iterations.

\section{Experimental Results}

This section briefly describes the details of training and presents the experimental results on the compared PDDL benchmark domains: Sokoban, Maze-with-Teleports, and Floor-Tile. A* algorithms with learnt heuristic functions realized by the proposed convolution-attention-position networks (further denoted as A*-CoAt) are compared to A* with learned heuristic function realized by convolutional networks as proposed in~\cite{groshev2017learning} (denoted as A*-CNN), and to the state of the art planners LAMA \cite{richter2010lama}, SymBA*~\cite{torralba2014symba}, and Mercury \cite{katz2014mercury}. We emphasize that A*-CNN and A*-CoAt uses vanilla A* search algorithm \cite{hart1968formal} without any additional tweaks. In case of Sokoban, we also compare our planner to a solution based on Reinforcement Learning~\cite{racaniere2017imagination}.

On all the compared domains, we analyse the strength of our learnt heuristic and generalization property by solving grid mazes of increasing complexities, approximated by the number of boxes in Sokoban, grids of higher dimensions in Floor-Tile and Maze-with-Teleports, and rotated mazes in Maze-with-Teleports.

\subsection{Training}
\paragraph{Sokoban:}
The policy-heuristic network we wish to learn accepts a state of a game, $s$ as an input and returns the next action, $a,$ and a heuristic value, $h(s),$ as an output. The training set $\xtrn = \{(s_i, a_i, |\pi^*(s_i)|)\}_{i=1}^n. $ therefore consists of $n \approx 10^6$ of these triples, i.e. $\xtrn = \{(s_i, a_i, |\pi^*(s_i)|)\}_{i=1}^n.$
The triples in the training set were created by randomly generating $40 000$ Sokoban instances using gym-sokoban \cite{SchraderSokoban2018}. Each instance has dimension $10 \times 10$ and it always contains only 3 boxes (and an agent). SymBA* \cite{torralba2014symba}, a planner that generates optimal solutions was used to generate optimal plans $\pi^*$ for each of these \textit{n} Sokoban instances. In each plan trajectory, the distance from a current state to the goal state is learned as the heuristic value, $h(s_i).$ Thus, the collection of all state-action-heuristic triples form the training set \xtrn.  


The Sokoban mazes in the \emph{training set were created with only three boxes}. This means that in the testing set, when we are solving instances with more boxes, we are evaluating its \emph{extrapolation} to more complex unseen environments, which cannot be solved by naive memorisation. However, our limited training set (containing 3 boxes) hinders the full potential of the neural network. With curriculum learning, we fine-tune the neural network using a training set containing Sokoban mazes of dimensions $10 \times 10$ with 3,4 5 6 and 7 boxes that have been already solved by the A*-NN with the corresponding architecture. This, therefore improves the heuristic function without the need to train the network from scratch and, more importantly, without the need to use other planners to create new plans.


\paragraph{Maze-with-Teleports:} The policy and heuristic network returns an action $a$, and a heuristic value, $h(s)$ as outputs. The training set \xtrn consists of $\approx$ 10000 maze problems of dimension $15 \times 15$. The mazes in \xtrn were generated using a maze creator\footnote{https://github.com/ravenkls/Maze-Generator-and-Solver} with slight modifications. Random walls were broken and a total of 4 pairs of teleports that connect different parts of the maze were created inside each training sample. As in the case of Sokoban, SymBA* \cite{torralba2014symba} was used to generate optimal solutions for the problems in the training set. The mazes for training were generated such that the initial position of the agent was in the upper-left corner and the goal was in the lower-right corner. Later,in our evaluations, we rotate each maze to investigate whether the heuristic function is rotation independent.

\paragraph{Floor-Tile:} The Floor-Tile heuristic network accepts a state of a game, $s$ as an input and returns a heuristic value, $h(s)$ as an output. The training set initially consists of $\approx$ 10000 Floor-Tile instances of dimension $4 \times 4$ with 2 agents. In our version of Floor-Tile, we assign \emph{white} to the first agent and \emph{black} to the second agent. The colors assigned to the agents are fixed and cannot be flipped at any stage of the game. SymBA* \cite{torralba2014symba} was used to generate optimal solutions for the Floor-Tile domain.

Since it is easy to generate Floor-Tile instances of increasing complexity (by increasing the size and varying the initial positions of the two agents), we experiment again with curriculum learning ~\cite{bengio2009curriculum} to develop scale-free heuristic values. Specifically, in our experiments we have quickly reached the capability of  planners at size $6 \times 5$ (more on this below). To further improve our heuristic function to scale to bigger problems, we re-train our network by extending the training set to include harder problem instances (of size $4\times 4$, $5 \times 5$, $6 \times 5$ and $6 \times 6$) the heuristic network has already solved.

%


All neural networks were trained by the Adam optimiser \cite{kingma2014adam} with a default learning rate of 0.001 for optimisation. In Sokoban and Maze-with-Teleports, the categorical cross entropy loss function was used to minimise the loss in the action prediction network and the mean absolute error loss was the loss function in the heuristic network. In Floor-Tile, the mean absolute loss was used in the heuristic network. For curriculum learning, domain, the learning rate was reduced to $ \approx 1 \times 10^{-4} $ in successive training iterations. Our experiments were conducted in Keras-2.4.3 framework with Tensorflow-2.3.1 as the backend. We used a NVIDIA Tesla GPU model V100-SXM2-32GB for training the neural networks.

\subsection{Comparison to prior State-of-the-Art} 

\paragraph{Sokoban:} The evaluation set consists of 2000 mazes of dimensions $10\times10$ with 3, 4, 5, 6 or 7 boxes (recall that the training set contain mazes with only 3 boxes).  Unless said otherwise, the quality of heuristics is measured by the relative number of solved mazes, which is also known as \textit{coverage}. Table \ref{tab:cvg} shows the coverage of compared planners, where \emph{all} planners were given 10 minutes to solve each Sokoban instance. We see that the classical planners solved all test mazes with three and four boxes but as the number of boxes increase, the A*-NN starts to have an edge. On problem instances with six and seven boxes, A*-CoAt achieved the best performance, even though it was trained only on mazes with three boxes. The same table shows, that A*-CoAt offers better coverage than A*-CNN, and we can also observe that curriculum learning (see column captioned curr.) significantly improves the coverage. 

We attribute SymBA*'s poor performance to its feature of always returning optimal plans while we are content with sub-optimal plans. LAMA had even lower success in solving more complicated mazes than SymBA*, despite having the option to output sub-optimal plans. To conclude, with an increase in the complexity of the mazes, the neural networks outshine the classical planners which makes them a useful alternative in the Sokoban domain.

The average plan length, shown in Table \ref{tab:planlen}, reveals that the heuristic learnt by the CoAt network is strong, as the average length of the plans is close to that of SymBA* which always returns optimal solutions.  We conclude that the proposed CoAt network delivers a strong heuristic outside its training, much better than that of the CNN~\cite{groshev2017learning} network and the planners (for mazes with more than 6 boxes).

\begin{table}[t]
\centering
\begin{tabular}{l>{\centering\arraybackslash}p{1.4em}>{\centering\arraybackslash}p{1.4em}>{\centering\arraybackslash}p{1.4em}>{\centering\arraybackslash}p{1.4em}>{\centering\arraybackslash}p{1.4em}>{\centering\arraybackslash}p{1.4em}>{\centering\arraybackslash}p{1.4em}>{\centering\arraybackslash}p{1.4em}>{\centering\arraybackslash}p{1.4em}} 
&&&&&\multicolumn{2}{c}{normal} & & \multicolumn{1}{c}{curr.} \\
\cline{6-7}
\cline{9-9}
\#b &  SBA* & Mrcy & LAMA && CNN &  CoAt && CoAt\\
\hline
3 &        \textbf{1} & \textbf{1} &      \textbf{1} &&    0.92  &   0.94 && 0.95\\ 
4 &        \textbf{1} & \textbf{1} &      \textbf{1} &&    0.87  &   0.91 && 0.93\\ 
5 &        0.95 & 0.75 &      0.89 && 0.83      &    0.89 && 0.91\\ 
6 &        0.69       &  0.60 &      0.65 && 0.69     &    \textbf{0.76} && \textbf{0.85}\\ 
7 &        0.45       &  0.24 &      0.32 && 0.58     &    \textbf{0.63} && \textbf{0.80}\\
\hline

\end{tabular}
\caption{Fraction of solved Sokoban mazes (coverage, higher is better) of SymBA* (SBA*), Mercury (Mrcy), LAMA,  A*-CNN (caption CNN) and the proposed A*-CoAt (caption CoAt). A*-CNN and A*-CoAt (with caption normal) use networks trained on mazes with three bozes; A*-CoAt (with caption curr.) used curriculum learning.}
\label{tab:cvg}
\end{table}{}

\begin{table}[t]
\centering
\begin{tabular}{lrrrrrr} 
\#b &  SBA* & Mrcy & CNN &  CoAt\\
\hline
3 &        21.40 & 21.70 &  24.20  & 22.20\\ 
4 &        34.00 & 34.33 &  40.53  & 36.00\\ 
5 &        38.82 & 42.83 &  45.52  & 39.11\\ 
6 &        41.11 & - &  51.00  & 42.11\\ 
7 &        -     & - &  54.33  & 44.17\\
\hline
\end{tabular}
\caption{Average plan length of SymBA*, Mercury, A* -CNN~\cite{groshev2017learning} (denoted as CNN) and that with the A*-CoAt. For clarity, we do not show results of LAMA, as it is performs exactly like SymBA* for 3 and 4 boxes. Column captioned \#b indicates the number of boxes in different categories.
}
\label{tab:planlen}
\end{table}

CoAt network is also on par with Deep Mind's implementation of Reinforcement Learning (DM-RL) in solving Sokoban \cite{racaniere2017imagination}. Instead of re-implementing DM-RL by ourselves, we report the results on their test set\footnote{Available at \url{https://github.com/deepmind/boxoban-levels}.} containing $10\times10$ Sokoban mazes with 4 boxes. While DM-RL had a coverage of 90\%, our A*-CoAt (trained on mazes with three boxes) has a coverage 87\%, and our A*-CoAt with curriculum learning has a coverage of 98.29\% \footnote{https://github.com/deepmind/boxoban-levels/blob/master/unfiltered/test/000.txt}. Taking into account that DM-RL's training set contained $10^{10}$ state-action pairs from mazes \textbf{with 4 boxes}, A*-CoAt achieves higher coverage using several orders of magnitude smaller training set.

\paragraph{Maze-with-Teleports:} The evaluation set contains a total of 2100 training samples of dimensions $15 \times 15$, $20 \times 20$, $30 \times 30$, $40 \times 40$, $50 \times 50$, $55 \times 55$ and $60 \times 60$. Each maze in the evaluation set contains 4 pairs of teleports that connects different parts of the maze. From Table \ref{tab:mtcoverage}, we see that the performance of A*-CNN and A*-CoAt (initially trained on $15 \times 15$ mazes) is the same as SymBA*\footnote{The planners and NNs were given 10 minutes to solve each maze instance.}  for dimensions up to $40 \times 40$ and is consistently better for problem instances of size $50 \times 50$, $55 \times 55$ and $60 \times 60$. 

All "No Rotation" mazes were created such that the agents start in the top left corner and the goal is in the bottom right corner. This allows us to study to which extent the learnt heuristic is rotation-independent (domain independent planners are rotation invariant by default). The same Table therefore reports fraction of solved mazes that have been rotated by 90$^{\circ}$, 180$^{\circ}$ and 270$^{\circ}.$ The results clearly show that the proposed heuristic function featuring CoAt blocks generalizes better than the one utilizing only convolutions, as the solved rotated instances of A*-CoAt network are comparable to the non-rotated case. Rotating mazes have no effect on SymBA* (the complexity is solely dependent on the grid size) and the coverage rate stays unaffected. 

From the results in Table \ref{tab:mtcoverage}, it can be concluded that the CoAt blocks (1) improve detection of non-local actions (teleports) compared to state-of-the-art planners such as SymBA*; (2) learn `useful' information from the mazes which makes the network robust to rotations; (3) learn to approximate distances inside the mazes which results in a scale-free heuristic function.
\begin{table*}[t]
\centering
\begin{tabular}{c| c c c| c c  c c | c c c |c c c } 
&&\multicolumn{2}{c}{No Rotation} & & \multicolumn{2}{c}{$90^\circ$ rotation} && \multicolumn{2}{c}{$180^\circ$ rotation} && \multicolumn{2}{c}{$270^\circ$ rotation} \\
size &          SBA* & CNN & CoAt &&  CNN & CoAt && CNN & CoAt  && CNN & CoAt \\
\hline
$15\times15$ &          1     & 1 &   1   &&   1 &  1   &&  1 & 1  && 1 & 1 \\ 
$20\times20$ &          1     & 1 &   1   &&   1 &  1   &&  1 & 1 && 1 & 1 \\ 
$30\times30$ &          1     & 1 &   1   &&   1 &  1   &&  1 & 1 && 1 & 1   \\ 
$40\times40$ &          1     & 1  &  1   &&   1 &  1   &&  1 & 1 && 1 & 1    \\ 
$50\times50$ &          0.92  & 0.94 & \textbf{1}  &&   0.91 & 1 &&  0.92 & 1  && 0.91 & 1\\
$55\times55$ &          0.55  & 0.78 & \textbf{0.89} && 0.71 & 0.85 && 0.70 & 0.87 && 0.69 & 0.87\\
$60\times60$ &          -     & 0.73 & \textbf{0.76} && 0.68 & 0.75 && 0.66 & 0.74 && 0.68 & 0.75\\
\hline

\end{tabular}
\caption{Fraction of solved mazes with teleports (coverage) of SymBA*, A* algorithm with convolution network~\cite{groshev2017learning} (denoted as CNN) and that with the proposed Convolution-Position-Attention (CoAt) network. Only non-rotated mazes (No Rotation) of size $15 \times 15$ were used to train the heuristic function. On mazes rotated by $90^\circ,$ $189^\circ,$ $270^\circ$, the heuristic function has to extrapolate outside its training set.}
\label{tab:mtcoverage}
\end{table*}

\paragraph{Floor-Tile:}
The evaluation set consists of 400 problem instances of sizes $5 \times 5$, $6 \times 5$, $6 \times 6$ and $7 \times 7$. Similarly to Sokoban, we first trained the CNN and CoAt NNs on small problem instances of size $4 \times 4$ (see columns denoted as "normal") and extrapolated it to instances of higher dimensions outside of the training set.
Table~\ref{tab:ftcov} shows the coverage of SymBA* and the heuristics learnt by the  NNs inside A* for different problem instances. The results are similar to those we have observed in Sokoban. On smaller problem instances, the classical planner SymBA* is better; on larger problem instances, the NNs are better.  The test set deliberately includes problem instances of size $6 \times 5$ to demonstrate the exact break point up to which SymBA* is able to generate solutions. Beyond grid size $6 \times 5$, \emph{all} state of the art planners fail and the solutions are generated only by the NNs. The heuristics generated by the A*-CoAt network can be extrapolated to solve 71\% of the tiling problems of size $6 \times 6$. On further increasing the grid dimension, the coverage of the A*-CoAt decreases to 29\% and even lower while A*-CNN is unable to generate solutions. The rightmost column (see column captioned ``curr'') in Table \ref{tab:ftcov} shows an improvement in coverage for dimensions $6 \times 6$ and $7 \times 7$ on implementing curriculum learning.


\begin{table}[t]
\centering
\begin{tabular}{lccccc} 
&&\multicolumn{2}{c}{normal} & & \multicolumn{1}{c}{curr.} \\
\cline{3-4}
\cline{6-6}
size            & SBA*          & CNN   & CoAt &  & CoAt   \\\hline
$4\times4$             &  \textbf{1}   & 0.92  &  0.96&  &  0.96     \\ 
$5\times5$             &   \textbf{1}  & 0.89  &  0.93&   & 0.93      \\ 
$6\times5$             &   \textbf{1}  & 0.78  &  0.88&   &  0.91     \\ 
$6\times6$             &     -         &  0.54 & \textbf{0.71}&  &  \textbf{0.89}  \\ 
$7\times7$             &     -         &  -    & \textbf{0.29}& & \textbf{0.76}  \\
 \hline
\end{tabular}
\caption{Fraction of solved Floor-Tile problem instances (coverage) of SymBA*, A*-CNN, and the proposed A*-CoAt. Fractions in columns captioned ``normal'' / ``curr.''  are of A* with heuristic functions trained on problem instances of size $4\times 4$ / by curriculum training respectively (see details in the text).}
\label{tab:ftcov}
\end{table}

\section{Conclusion and Future Work}

We showed that learning a strong heuristic function for PDDL domains with underlying grid structure is possible without the need for any specific domain knowledge. In fact, the architecture of the learning approach is general to any PDDL-type planning problem, and the learning can be executed on any vector of propositions, or finite-domain variables as states forming the input to the neural network. Identifying some structure in advance is advantageous.

While the heuristic function can generate sub-optimal plans, our experiments suggest that the plan quality is not far from the optimum. Moreover, while we have generated training data from a classical planner on small problem sizes, the proposed architecture is able to generalize and successfully solve more difficult problem instances, where it surpasses classical domain-independent planners, while improving on previously known state-of-the-art. 

Our experiments further suggest that the learnt heuristic can further improve, if it is retrained / fine-tuned on problem instances it has previously solved. This form of curriculum learning aids the heuristic function in solving mainly large and more complex problem instances that are otherwise not solvable by domain independent planners within 10 minutes.

As future work, our next goal would be to better understand if the learnt heuristic function is similar to something that is already known, or something so novel that it can further enrich the field; i.e., what kind of underlying problem structure we can learn by which network type best, possibly in form 
of studying generic types~\cite{LongF00}. 

Our experiments with curriculum learning suggest that the neural network might boot-strap itself by first learning on simple trivial examples and gradually solving more difficult ones by adding them to the training set. This raises two questions, mainly regarding the limit of this process and if the gains obtained by using larger networks will vanish. 

We  believe that an improvement in the heuristic function is tied to the generation of problem instances that inherently possess the right level of difficulty, by which we mean that they have to be just on the edge of solvability, such that the plan can be created and added to the training set. We are fully aware that the problem instance generation itself is a hard problem, but we cannot imagine the above solution to be better than specialized 
domain-dependent Sokoban solvers without such a generator (unless the collection of all Sokoban mazes posses this property).
We also question the average estimation error minimized during learning of the heuristic function. It might put too much emphasis on simple problem instances that are already abundant in the training set while neglecting the difficult ones.
We wish to answer some of the above question in the future in an endeavour to generate strong, scale-free heuristics.

\vfill

\pagebreak

\begin{small}
\bibliography{aaai22.bib}
\end{small}

\end{document}